\documentclass[10pt,twocolumn,letterpaper]{article}

\usepackage{cvpr}
\usepackage{times}
\usepackage{epsfig}
\usepackage{graphicx}
\usepackage{amsmath}
\usepackage{amssymb}

\usepackage{booktabs}
\usepackage{bm}
\usepackage{color}
\usepackage{amsmath}
\usepackage{amssymb}
\usepackage{subfigure}
\usepackage{color}
\usepackage{multirow}
\usepackage[breaklinks=true,bookmarks=false]{hyperref}

\cvprfinalcopy % *** Uncomment this line for the final submission

\def\cvprPaperID{****} % *** Enter the CVPR Paper ID here

\begin{document}

%%%%%%%%% TITLE
\title{Unsupervised Multi-Domain Image Translation with Domain-Specific Encoders/Decoders}

\author{Le Hui ~ ~ Xiang Li ~ ~ Jiaxin Chen ~ ~ Hongliang He ~ ~ Chen Gong ~ ~ Jian Yang \\
Nanjing University of Science and Technology \\
 \tt{\small \{le.hui, xiang.li.implus, jiaxinchen, HongliangHe, chen.gong, csjyang\}@njust.edu.cn}}

%\author{Le Hui\\
%Nanjing University of Science and Technology\\
%Institution1 address\\
%{\tt\small firstauthor@i1.org}
%% For a paper whose authors are all at the same institution,
%% omit the following lines up until the closing ``}''.
%% Additional authors and addresses can be added with ``\and'',
%% just like the second author.
%% To save space, use either the email address or home page, not both
%\and
%Xiang Li\\
%Institution2\\
%First line of institution2 address\\
%{\tt\small secondauthor@i2.org}
%\and
%Jiaxin Chen\\
%Institution2\\
%First line of institution2 address\\
%{\tt\small secondauthor@i2.org}
%\and
%Hongliang He\\
%Institution2\\
%First line of institution2 address\\
%{\tt\small secondauthor@i2.org}
%\and
%Chen Gong\\
%Institution2\\
%First line of institution2 address\\
%{\tt\small secondauthor@i2.org}
%\and
%Jian Yang\\
%Institution2\\
%First line of institution2 address\\
%{\tt\small secondauthor@i2.org}
%}

\maketitle

\begin{figure*}[t]
	\begin{center}
		\setlength{\fboxrule}{0pt}
		\fbox{\includegraphics[width=\textwidth]{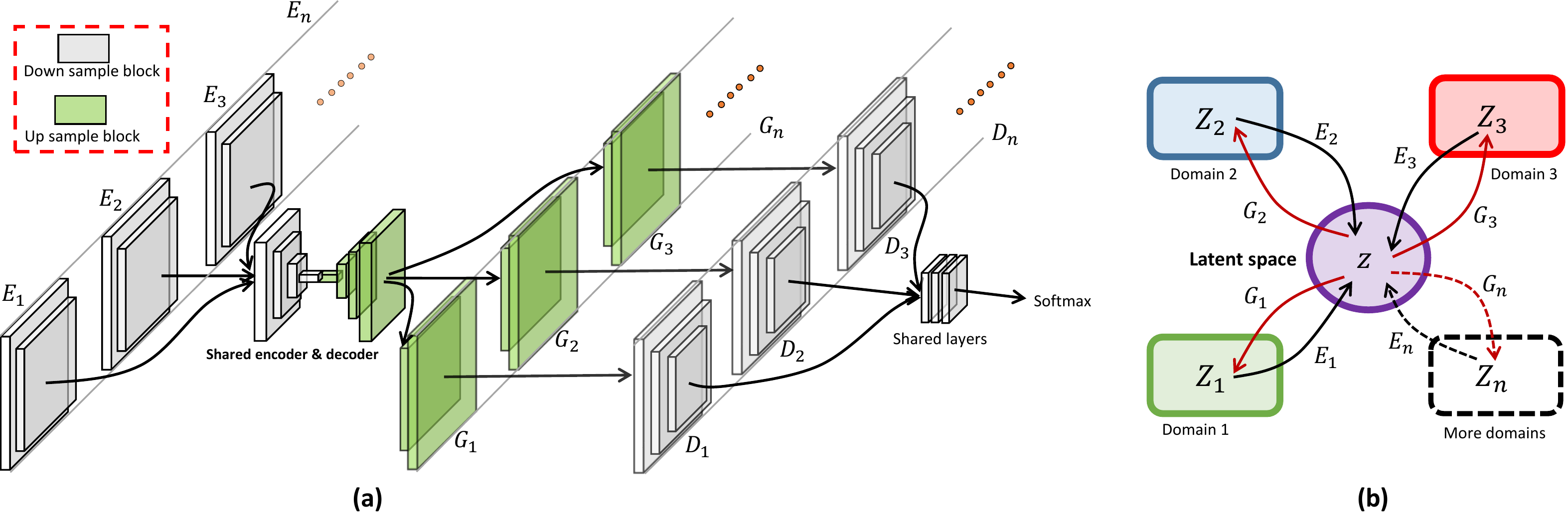}}
	\end{center}
	\caption{(a) The proposed Domain-Bank framework. We declare tuples of $\{E_1, ..., E_n\}$ and $\{G_1, ..., G_n\}$. By adopting a weight sharing constraint in the last few layers of $\{E_1, ..., E_n\}$ and the first few layers of $\{G_1, ..., G_n\}$, we implement the shared-latent space assumption. (b) The shared latent space assumption in \cite{liu2017unsupervised}. For arbitrary pairs of corresponding image ($x_i^a, x_j^b$), ($a, b \in [1, n]$, $n$ is the number of domains), they can be translated to a same latent code $z$. $\{E_1, ..., E_n\}$ are encoding functions that map image to a latent code, and then $\{G_1, ..., G_n\}$ are decoding the latent code to images of corresponding domain. $\{D_1, ..., D_n\}$ are adversarial discriminators for the corresponding domains in order to evaluate whether the translated images are realistic.}
	\vspace{-15pt}
	\label{fig_model_cropped}
\end{figure*}

%%%%%%%%% ABSTRACT
\begin{abstract}
   Unsupervised Image-to-Image Translation achieves spectacularly advanced developments nowadays. However, recent approaches mainly focus on one model with two domains, which may face heavy burdens with large cost of $O(n^2)$ training time and model parameters, under such a requirement that $n$ domains are freely transferred to each other in a general setting. To address this problem, we propose a novel and unified framework named Domain-Bank, which consists of a global shared auto-encoder and $n$ domain-specific encoders/decoders, assuming that a universal shared-latent sapce can be projected. Thus, we yield $O(n)$ complexity in model parameters along with a huge reduction of the time budgets. Besides the high efficiency, we show the comparable (or even better) image translation results over state-of-the-arts on various challenging unsupervised image translation tasks, including face image translation, fashion-clothes translation and painting style translation. We also apply the proposed framework to domain adaptation and achieve state-of-the-art performance on digit benchmark datasets. Further, thanks to the explicit representation of the domain-specific decoders as well as the universal shared-latent space, it also enables us to conduct incremental learning to add a new domain encoder/decoder. Linear combination of different domains' representations is also obtained by fusing the corresponding decoders.
\end{abstract}

%%%%%%%%% BODY TEXT
\section{Introduction}
Image-to-image translation problem is a general formulation which involves a wide range of various computer vision problems. Just as a sentence may be translated in either English or French, an image may be rendered in another image. Many problem in image processing can be defined as ``translating'' an input image in one domain into a corresponding output image in another domain. Typically, denoising, super-resolution and colorization all pertain to image-to-image translation where input is a degraded image (noisy, low-resolution, or gray scale) and the output is a high-quality color image.

%应用 and 算法
Recently, a series of attractive works ignite a renewed interest in the image-to-image translation problem by adopting Convolution Neural Networks (CNNs). Gatys et al. \cite{gatys2015neural} first study how to use CNN to reproduce famous painting styles on natural images. Since the seminal work by Goodfellow et al. \cite{goodfellow2014generative}, GAN has been proposed for a wide variety of problems. Unlike past works, by utilizing GANs, \cite{liu2017unsupervised,liu2016coupled,yi2017dualgan,zhu2017unpaired} are proposed to translate an image from a source domain $X$ to a target domain $Y$ in the absence of paired examples. These algorithms often produce more impressive results near to the corresponding target domain, since a joint distribution that can be learnt from two different domains by using images from the marginal distributions in individual domains.

Notwithstanding their demonstrated success, currently existing approaches basically focus on the one model with two domains setting. Specially, learnt through one fresh training, translation is limited to transfers one pair of different domains. After a careful examination of existing image-to-image translation networks, we argue that different marginal distributions can be projected into a common space in their learnt network structures. To the best of our knowledge, a translation among $n$ domains has not yet been proposed with $O(n)$ complexity in these previous works.

As a result, the network is only able to capture a two specific domains translate one at a time. For a new domain, the whole network has to be retrained end-to-end, which leads to an unavoidable burden under the situation where $n\times(n-1)$ transformations are required, given $n$ domains. In practice, this make these methods unable to scale to a large number of domains, especially when the domain require to be incrementally augmented. Additionally, how to further reduce the training time, network model size and enable more flexibilities to control translation among domains, remain to be considered yet to be addressed.

To overcome these problems, we explore a multi-domain image translation in which we reconsider the joint distributions of multiple domains. From the perspective of a probabilistic modeling, the coupling theory \cite{lindvall2002lectures} states there exists an infinite set of joint distributions that can arrive the given marginal distributions in general. This highly ill-posed problem forces us to make additional assumption on the structure of the joint distribution. By further considering the interaction of $n$ domains, we make a global shared-latent space assumption that assumes every sampled image from one of $n$ domains can be mapped to an universal shared-latent space. Based on the universal assumption, we propose a compact, and easily extended \emph{Domain-Bank} framework that learns every domain pairs' joint distribution simultaneously.

In details, the proposed \emph{Domain-Bank} framework is composed of multiple domain-specific component banks and each component represents one specific domain. Specially, a component bank consist of encoder and decoder for a specific domain. For an input image, the corresponding component bank maps an image to a shared-latent space, and then decodes it to the target image.

%a code in a shared-latent space via the encoder, and then decodes a random-perturbed version of the code to generate the target image. Therefore, we have achieved the translation of images from one domain to another.

In several challenging unsupervised multi-domain image translation tasks like face image translation, fashion-clothes translation and painting style translation, we comprehensively demonstrate the superior efficiency and at least comparable results to the state-of-the-art methods. Furthermore, as more domains' samples are engaged in \emph{Domain-Bank} framework, the performance gain of domain adaptation tasks on digital recognition becomes consistently obvious. More importantly, it not only allows us to simultaneously learn the translation among various domains, but also enables a very efficient incremental learning for a new image domain. This is achieved by learning a new component bank of domain while holding the other auto-encoder/decoder fixed.

Compared with existing models under the unsupervised image-to-image translation settings, our proposed \emph{Domain-Bank} is unique in the following aspects:

\begin{itemize}
	\item Our model is designed with a compact and clean structure, which also obtains a considerably huge reduction (from a complexity of $O(n^2)$ to $O(n)$) of training time and model parameters in case of $n$ ($n > 2$) domains with $n \times (n - 1)$ transformations.
	\item The universal shared auto-encoder subnetwork is trained efficiently and effectively with multi-domain training samples/pairs, thus leading to a better generation which is confirmed in both quantitative and qualitative experimental results.
	\item The shared auto-encoder, along with the domain-specific encoders/decoders, can provide more functional utilizations like domain linear combination or incrementally learning a new domain.
\end{itemize}

The remainder of the paper is organized as follows: Related work is summarized in Section \ref{sec_related_work}. We devote Section \ref{sec_unitbank} to the main technical design of the proposed \emph{Domain-Bank}. Section \ref{sec_exp} gives the experimental results in both quantitative and qualitative aspects. New characteristics of the proposed framework can be found in Section \ref{sec_capability}. Finally, we conclude our paper in Section \ref{sec_conclusion}.

%-------------------------------------------------------------------------
\section{Related Work}
\label{sec_related_work}
Image-to-Image translation problem has already been promoted by deep neural network and obtains some impressive results especially in the domain/style transfer fields. Neural generative models has recently received an increasing amount of attention. Several algorithms, including generative adversarial networks \cite{goodfellow2014generative}, variational autoencoders (VAEs) \cite{kingma2013auto,larsen2015autoencoding}, stochastic back-propagation \cite{rezende2014stochastic} and diffusion processes\cite{sohl2015deep}, have demonstrated that a deep neural network can learn a domain distribution from examples. Thus, the learned networks can be used to generate novel images. We are interested in image-to-image translation problem. After analyzing the image translation problem from a probabilistic modeling attitude, the key challenge is to learn a joint distribution of images in different domains from its marginal distributions of individual domains.

\textbf{Image-to-Image Translation.} Many image processing and computer vision tasks can be posed as an image-to-image translation problem, mapping an image in one domain to a corresponding image in another domain, e.g., image segmentation, stylization, super-resolution and abstraction. Particularly, the image-to-image can be traced back at least to Hertzmann et al's Image Analogies \cite{hertzmann2001image}. Hence, image segmentation can be considered as a problem of mapping a natural image to a corresponding segmented image. More recent approaches use a dataset of input-output samples to learn a parametric function using CNNs. Similar ideas have also been applied to various tasks including generating photo from sketches or attributes and semantic layouts etc.

\textbf{Unsupervised Image-to-Image Translation.} In unsupervised image-to-image setting, we only have two independent sets of images where one possesses images in one domain and so do the other. Note that there exists no paired samples guiding how an image could be translated to a corresponding image in another domain. Several other approaches also adopt the unpaired setting, where the goal is to relate two data domains, domain $X$ and domain $Y$. More Recently, \cite{taigman2016unsupervised} proposed the domain transformation network (DTN) and achieved promising results on translating small resolution face and digit images. Liu et al. proposed CoGAN \cite{liu2016coupled}, which use a weight-sharing strategy to learn a common representation across two domains. Following, Liu et al. first made a shared-latent assumption, and then they proposed an unsupervised image-to-image translation framework \cite{liu2017unsupervised}, which uses a VAEs and GANs to learn a mapping from input to output images. Our approach builds on the this framework. However, unlike these prior works, we learn the translation among multiple domains (more than two domains) without paired training samples and also enable a very efficient incremental learning for a new domain based on our proposed framework at the same time.

\textbf{Generative Adversarial Networks (GANs).} GANs have achieved great success in a wide variety of computer vision applications, enhancing both supervised tasks and unsupervised ones. The key of GANs is the introduction of the \emph{adversarial loss}, that forces the generated images to be indistinguishable from real images substantially. Learning in GAN is via staging a zero-sum game between two players, where the discriminator tries to distinguish reliable real samples from fake ones and the generator attempts to fool it. Soon after, various GANs have been proposed to the image generation on class labels \cite{mirza2014conditional}, attributes \cite{perarnau2016invertible,yan2016attribute2image} and images \cite{ledig2016photo,liu2017unsupervised,zhu2017unpaired,liu2016coupled,yi2017dualgan,radford2015unsupervised,isola2016image}. A list of training tricks of GANs is given in \cite{salimans2016improved}.

\textbf{Variational Auto Encoders (VAEs).} A VAE consists of two networks that encode a data sample to a latent representation and decode the latent representation back to data space. The key of VAEs is to optimize a variational bound. By enhancing the variational approximation, superior image generation results were obtained \cite{kingma2016improving,maaloe2016auxiliary}. Larsen et al. \cite{larsen2015autoencoding} proposed a VAE-GAN architecture to improve image generation quality of VAEs. VAEs also were applied to translate face image attribute in \cite{yan2016attribute2image}. More recently, Liu et al. \cite{liu2017unsupervised} extend the framework of VAE-GAN to unsupervised image-to-image translation problems.

\section{Domain-Bank Networks}
\label{sec_unitbank}
Our goal is to learn $n \times (n - 1)$ translations for $n$ domains. It may offer a new understanding for the image domain translation problems, and then help design a more elegant architecture to address multi-domain translation problems.

We construct a multi-domain image translation network based on variational autoencoders (VAEs) \cite{kingma2013auto,larsen2015autoencoding,rezende2014stochastic} and generative adversarial networks (GANs) \cite{goodfellow2014generative,liu2016coupled,zhu2017unpaired}, which encodes an input image to the shared-latent space and can also reconstruct/transfer it.

%Certainly, our \emph{Domain-Bank} embodies multiple domain-specific component banks. Each component bank contains a specific-encoder and a specific-decoder. By decoding the shared-latent code, produced by specific-encoder, \emph{Domain-Bank} would map the source images to the corresponding content images.
\begin{table*}
	\centering
	\begin{tabular}{c|ccccc}
		\toprule
		{\bf Networks} & \{$E_a, G_a$\} & \{$G_a, D_a$\} & \{$E_a, G_a, D_a$\} & \{$E_a, G_b$\} & \{$E_a, G_b, E_b, G_a$\}\\
		\midrule
		{\bf Functions} & VAE for $X_a$ & GAN for $X_a$ & VAE-GAN \cite{larsen2015autoencoding} & Image Translator $X_a \rightarrow X_b$ & Cycle-consistency \cite{zhu2017unpaired}\\
		\bottomrule
	\end{tabular}
	\caption{Interpretation of the functions of network architecture.}
	\vspace{-15pt}
	\label{tab:network_roles}
\end{table*}
\subsection{Network Architecture}
Figure \ref{fig_model_cropped} shows our multi-domain translation architecture, which is based on the universal shared-latent space assumption. Suppose we are considering arbitrary two domains of $n$ domains, namely $X_a, X_b, a, b \in [1, n], a \ne b$, which contain training samples $\{x_i^a\}_{i=1}^{N_a}$ where $x_i^a \in X_a$ and $\{x_j^b\}_{j=1}^{N_b}$ where $x_j^b \in X_b$, respectively. We denote the corresponding marginal data distribution as $x^a \sim P_{X_a}$ and $x^b \sim P_{X_b}$. We aim to learn a joint distribution of images in domain $X_a$ and $X_b$ by utilizing images from the marginal distributions in two individual domains. It can be easily extended to the case of $n$ domains. That is, we can learn a joint distribution of $n$ domains.

Every one (supposed to be $a, a \in [1, n]$) has two functional paths, through which any given image $x^a$ sampled from $P_{X_a}$ can be projected into the shared-latent code $z$ and it can be recovered back as well. That is, we suppose there exists functions $E_a, E_b, G_a, $ and $G_b$ ($a, b \in [1, n]$) such that, given a pair of corresponding images ($x_i^a, x_j^b$) (where $x_i^a \in X_a$, $x_j^b \in X_b$, $i \in [1, N_a]$ and $j \in [1, N_b]$) from the joint distribution. We define $z = E_a(x_i^a) = E_b(x_j^b)$, on the contrary, $x_i^a = G_a(z)$ and $x_j^b = G_b(z)$. In our structures, we map domain $X_a$ to domain $X_b$ through the function $x_j^b = F_{a \rightarrow b}(x_i^a)$, which can be represented by the function $F_{a \rightarrow b}(x_i^a) = G_b(E_a(x_i^a))$. Equally, we define two reconstruction functions for domain $X_a$ to domain $X_a$: 1$)$ $x_i^a = F_{a \rightarrow a}(x_i^a)$ and 2$)$ $ x_i^a = F_{a \rightarrow b \rightarrow a}(x_i^a)$. The function 1$)$ can be equivalently written as $F_{a \rightarrow a}(x_i^a) = G_a(E_a(x_i^a))$, and the 2$)$ is written as $F_{a \rightarrow b \rightarrow a}(x_i^a) = G_a(E_b(F_{a \rightarrow b}(x_i^a))) = G_a(E_b(G_b(E_a(x_i^a))))$. More notably, for an input image in domain $X_a$, the function 1$)$ directly translate it to an image in domain $X_a$. However, in function 2$)$, the input images are first translated from domain $X_a$ to domain $X_b$, and then the generated images in domain $X_b$ are converted back to the domain $X_a$. In addition, a necessary condition for translating domain $X_a$ to domain $X_b$ to exist is the cycle-consistency constraint \cite{kim2017learning,liu2017unsupervised,zhu2017unpaired}: $F_{a \rightarrow b \rightarrow a}(x_i^a) = F_{b \rightarrow a}(F_{a \rightarrow b}(x_i^a))$. In other words, we can reconstruct the input image from translating back to the translated input image. Therefore, the shared-latent space assumption indicates the cycle-consistency assumption.

\textbf{Domain-Specific Encoder and Decoder.} Following the architecture used in \cite{liu2017unsupervised}, the image encoder $E_a$ consists of 3 convolutional layers and 3 basic residual blocks \cite{he2016deep}, symmetrically, the image decoder $G_a$ also consists of 3 basic residual blocks and 3 transposed convolutional layers. In our mulit-domain image-to-image translation, different domains have domain-specific encoders $E_a$ and domain-specific decoders $G_a$. For instance, Monet's painting need to use Monet's specific encoder whilst Van Gogh's has to use Van Gogh's specific encoder. Similarly, this is also necessary for the domain-specific decoders. Different domains use domain-specific encoders to extract representations of the input images, and then domain-specific decoders responsible for decoding representations for reconstructing images in different domains. In other words, the encoder and the decoder can be seen as a domain-specific component, and we only need to train different components for different domains. In practice, when a new domain arrives, it also enables us to conduct incremental learning to train the encoder and decoder.

%\noindent%
\textbf{Universal Shared Auto-Encoder.} Based on the shared-latent assumption, we enforce a weight-sharing constraint to relate the VAEs. Specially, we further assume a share intermediate representation $h$ that the process of generating corresponding images satisfy the formula
\vspace{-5pt}
\begin{equation}
\{x^a, x^b, ..., x^n\} \rightarrow h \rightarrow z.
\vspace{-5pt}
\end{equation}

Therefore, we assume $E_a = E_{L, a} \circ E_H$ where $E_{L, a}$ is a common low-level generation function that maps $X_a, a \in [1, n]$ to $h$, respectively. However, $E_{H}$ are high-level generation function that maps $h$ to $z$. From another view , $z$ can be considered as the high-level representation of different domains, and $h$ can be regarded as a special implementation of $z$ through $E_H$. Similarly, $h$ also admit us to represent $G_a$ by $G_a = G_H \circ G_{L, a}.$ In the implementation, we share the weights of the last few layers of $E_a, a \in [1, n]$ that are responsible for extracting high-level representations of input images in the $n$ domains. Equally, the first few layers of $G_a, a \in [1, n]$ are shared, which responsible for decoding high-level representations for reconstructing the input images.

\textbf{Domain-Specific Discriminator.} Since we have $n$ different domains, our framework has $n$ adversarial networks: $GAN_a$ = \{$D_a, G_a$\}. In $GAN_a$, for real images sampled from the domain $X_a$, $D_a$ should output true, while for images generated by $G_a$, it should output false. In our framework,  $GAN_a$ can generate two types of images: 1$)$ images from the reconstruction streams $F_{a \rightarrow a} = G_a(E_a(z))$ and 2$)$ images from the translation streams $F_{a \rightarrow b} = G_b(E_a(z))$. The reconstruction streams can be trained with supervisions that we only apply adversarial training to images from the translation streams, $F_{a \rightarrow b}$. Thus, we require train $n$ domain-specific discriminators for $n$ different domains.

\subsection{Loss Function}
To better understand the losses applied in \emph{Domain-Bank}, we first give a decomposed perspective of possible combinations of the key components in Figure \ref{fig_model_cropped}. Basically, our framework is based on variational autoencoders (VAEs) and generative adversarial networks (GANs) including $n$ domain image encoders $E_a$, $n$ domain image generators $G_a$ and $n$ domain adversarial discriminators $D_a$ where $a \in [1, n]$. The Table \ref{tab:network_roles} further explains the various roles inside our framework and their corresponding functions.

In the image-to-image translation problem of our \emph{Domain-Bank}, we have three kinds of fundamental information streams, namely the image reconstruction streams, the image translation streams, and the cycle-reconstruction streams. In Table \ref{tab:network_roles}, the encoder-decoder pair \{$E_a, G_a$\} constitutes VAE for the image reconstruction streams. For an input domain $X_a$, the image is translated to another domain $X_b$ by the translation stream \{$E_a, G_b$\}. Since the shared-latent space assumption indicates the cycle-consistency constraint, we require a cycle-reconstruction stream \{$E_a, G_b, E_a, G_a$\} to reconstruct input images. Consequently, we are jointly considering address the problem of VAE and GAN to solve the image translation problem.

\textbf{VAE loss.} In our framework, we use variational autoencoder (VAE) to generate images in which VAE is supervised by the $\mathbf{KL}$ divergence. In the VAE, the encoder outputs a mean vector $E_{\mu, a}(x_i^a)$ where the input image $x_i^a \in X_a$.  The distribution of the latent code $z$ is written as $q_a(z_a | x_i^a) \equiv \mathcal{N}(z_a|E_{\mu,a}, I)$ where $I$ is an identity matrix. We assume the distribution of $q_a(z_a | x_i^a)$ as a random vector of $\mathcal{N}(z_a|E_{\mu,a}, I)$ and sample from it. Thus, the reconstructed image is $ x_{i}^{a \rightarrow a} = G_a(z_a \sim q_a(z_a | x_i^a))$. In addition, let $\eta$ be a random vector with a multi-variate Gaussian distribution: $\eta \sim \mathcal{N}(\eta|0, I)$. In the VAE, the function $z_a \sim q_a(z_a|x_i^a)$ is implemented via $z_a = E_{a, \mu}(x_i^a) + \eta$. The aim of VAE is to minimize a variational upper bound that the VAE object is written as
\begin{eqnarray}
\mathcal{L}_{VAE_a}(E_a, G_a) = \lambda_1 \mathbf{KL}(q_a(z_a|x^a) \| p_{\eta}(z)) \nonumber \\
- \lambda_2\mathbb{E}_{z_a \sim q_a(z_a | x^a)} \left[ \log p_{G_a}(x^a | z_a) \right],
\end{eqnarray}
where $\lambda_1$ and $\lambda_2$ display the weights of corresponding objective and the $\mathbf{KL}$ divergence term penalizes the deviation of the latent code distribution from the prior distribution. More notable, the L1 loss used in VAE is to ensure similarity between the generated real image and the original rendering image.

\textbf{Adversarial loss.} The adversarial losses are applied to translation functions. Note that we have defined $G_a$ and its discriminator $D_a$, where $a \in [1, n]$. We express the objective as:
\begin{eqnarray}
\mathcal{L}_{GAN_{ab}}(E_a, G_a, D_a) = \lambda_0 \mathbb{E}_{x^a \sim P_{x^a}} \left[ \log D_a(x^a) \right]\nonumber \\
+ \lambda_0\mathbb{E}_{z_b \sim q_b\left( z_b | x^b \right)} \left[ \log \left( 1 - D_a(G_a(z_b)) \right) \right],
\end{eqnarray}
where the hyper-parameter $\lambda_0$ controls the impact of the GAN objective functions. In the adversarial part, $G_a$ attempts to generate images $G_a(z_b)$ that look like images from domain $X_a$, while $D_a$ tries to distinguish between translated samples $G_a(z_b)$ and real samples $x^a$. Finally, $G_a$ aims to minimize this objective against an adversary $D_a$ that tries to maximize it.

\textbf{Cycle-consistency Loss.} We use a VAE-like function to model the cycle-consistency constraint, which is written as
\begin{eqnarray}
\mathcal{L}_{cyc_{ab}}(E_a, G_a, E_b, G_b) = \lambda_3\mathbf{KL}(q_a(z_a|x^a) \| p_{\eta}(z)) \nonumber \\
+ \lambda_3\mathbf{KL}(q_b(z_b|F_{a \rightarrow b}(x^a)) \| p_{\eta}(z)) \nonumber \\
- \lambda_4\mathbb{E}_{z_b \sim q_b(z_b | F_{a \rightarrow b}(x^a))} \left[ \log p_{G_a}(x^a | z_b)  \right].
\end{eqnarray}

\textbf{Full objective.}
As a result, our ultimate objective is written as:
\vspace{-10pt}
\begin{eqnarray}
\mathcal{L}(E, G, D)=\sum_{a}^{n} \sum_{b}^{n} \{{\mathcal{L}}_{VAE_a}(E_a, G_a) \nonumber \\
+ \mathcal{L}_{GAN_{ab}}(E_a, G_b, D_b) \nonumber \\
+ \mathcal{L}_{cyc_{ab}}(E_a, G_b, E_b, G_a)\},
\end{eqnarray}
where $a, b \in [1, n]$ and $a \neq b$. We aim to solve:
\begin{equation}
E^*,G^* = \arg\mathop{\min}_{E, G} \mathop{\max}_{D} \; \mathcal{L}(E, G, D).
\end{equation}

\begin{figure*}[t]
	\vspace{-2pt}
	\begin{center}
		\setlength{\fboxrule}{0pt}
		\fbox{\includegraphics[width=\textwidth]{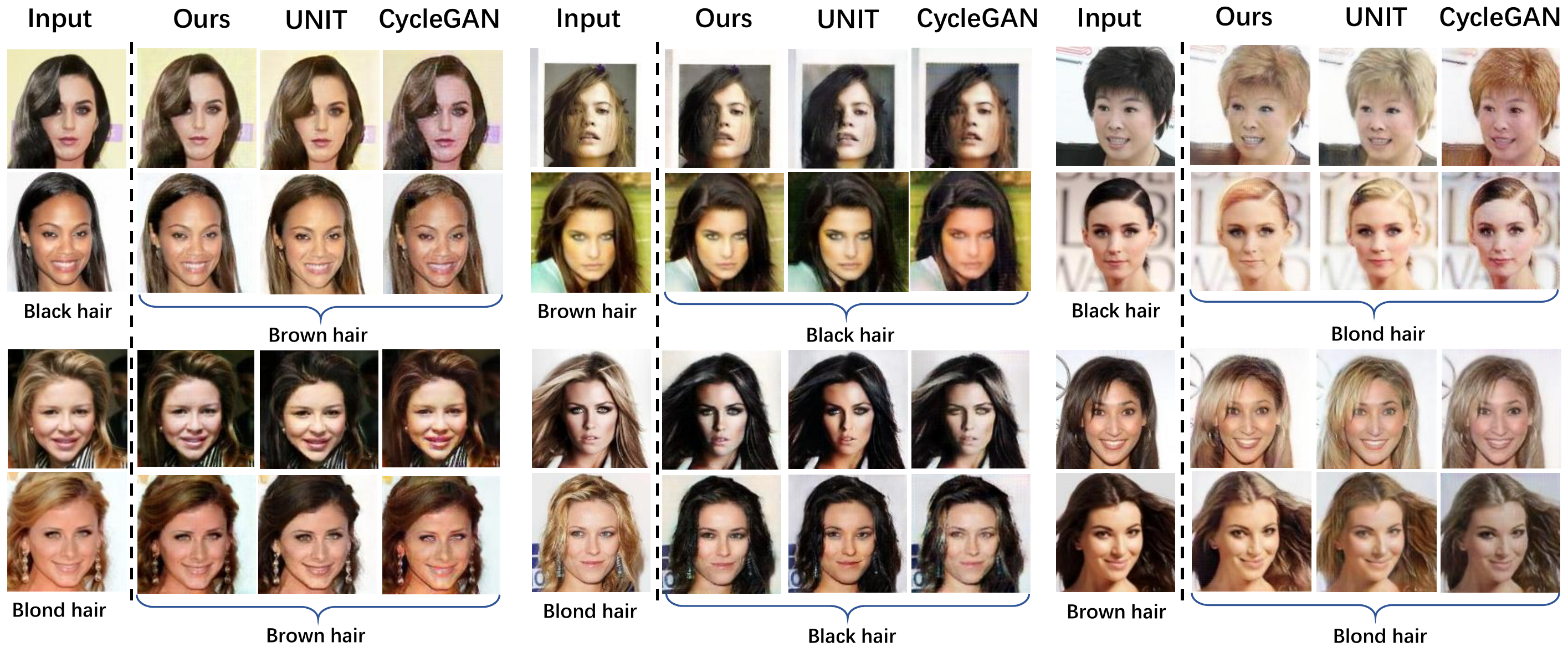}}
	\end{center}
	\vspace{-10pt}
	\caption{The results of Face attributes translation. It contains the results of three sets of hair translation. In each group, the far left is the input image, and the results from left to right are Ours, UNIT and CycleGAN. We obtain comparable results with others. In particularly, our method better preserves the original quality of human face and facial identity. It can be observed that only the color of hair region changes, and the rest remains even in details. In addition, our results are generated by passing end-to-end training only once, while the others require training three models for translations between different pairs of domains respectively.}
	\label{fig:face}
\end{figure*}
\begin{figure*}[t]
	\vspace{-10pt}
	\begin{center}
		\setlength{\fboxrule}{0pt}
		\fbox{\includegraphics[width=\textwidth]{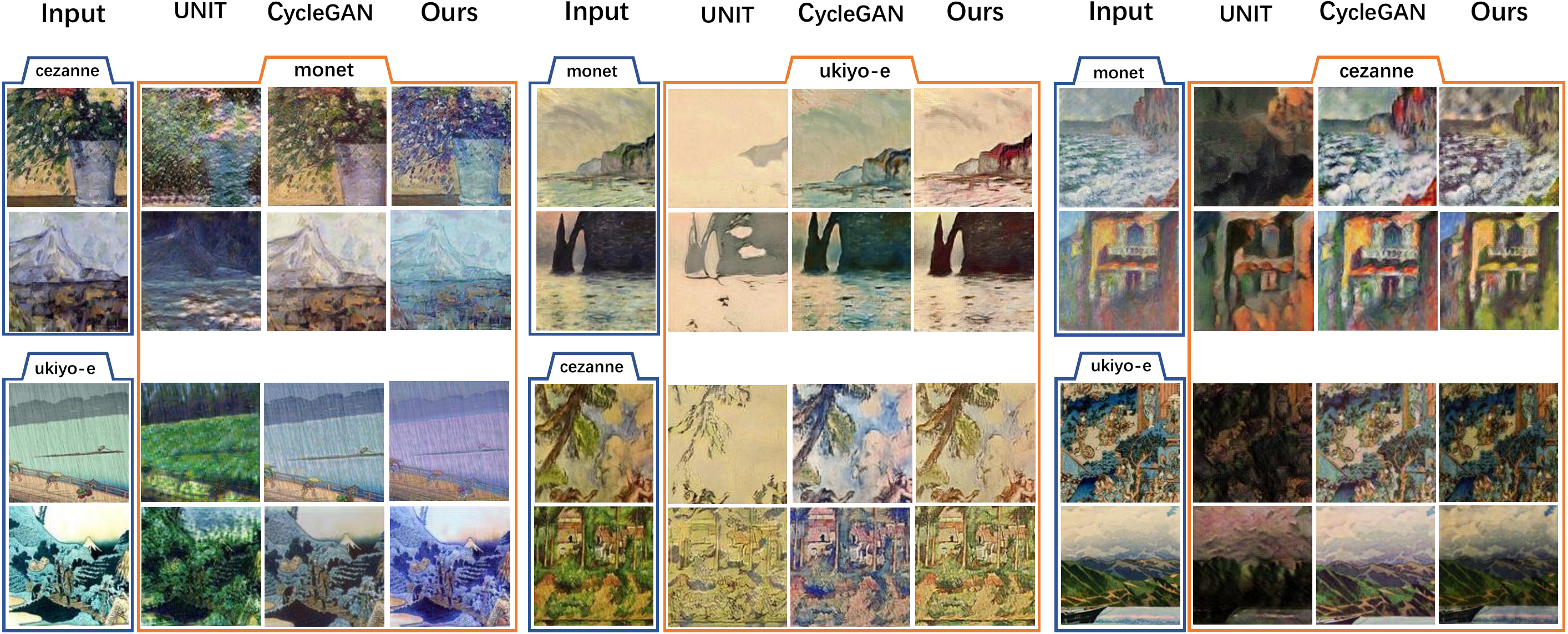}}
	\end{center}
	\vspace{-10pt}
	\caption{The results show the painting style translation of famous painters: Cezanne, Monet, and Ukiyo-e. Our generated images are more clear and with higher contrast ratio than UNIT's, and can be near the target domain in the aspect of style. All the results we have obtained are visibly superior than UNIT's. On the image quality, our generated images also have high contrast when compare to CycleGAN. Note that all our results are obtained by training only once avoiding heavy training burden while others not.}
	\vspace{-10pt}
	\label{fig:painting}
\end{figure*}

\subsection{Training Strategy}
We employ an alternative training strategy motivated by GAN's \cite{goodfellow2014generative} solving a mini-max problem where the optimization aims to find a saddle point. The zero-sum game in our framework consists of two plays: the domain-specific discriminators as the first team, and the domain-specific encoders/decoders for the second. During training with a specific pair of images from domain $X_a$ and $X_b$, we first train domain-specific $X_b$'s discriminator with all other components fixed. Afterwards the $X_a$'s encoder/decoder and $X_b$'s encoder/decoder are involved not only to minimize the VAEs losses and the cycle-consistency losses but also to defeat the first player.

\section{Experiments}
\label{sec_exp}
We first give qualitative results on various tasks along with rich complexity comparisons. Further, we present the quantitative performance gain on the digital domain adaptation tasks.
\subsection{Qualitative Analysis}

\noindent \textbf{Face attributes translation.} The CelebA dataset \cite{liu2015deep} is exploited for attribute-based face images translation. There are many different attributes of face images including hair, smiling and eyeglass. Particularly, we select a domain of hair with different colors including blond, brown, black, etc. Specifically, the hair with blond color constitutes the 1st domain, the brown hair constitutes the 2nd domain, while the black hair constitutes the 3rd domain. In Figure \ref{fig:face}, we visualize the results where we display the transitions between hair with different colors. We find that the translated hair images are impressive. It is not difficult to see that we obtain comparable results to other algorithms in hair translation.

When training for multi-domain translation, the shared-latent space of our framework accurately captures the invisible inner-similarity whereas encoders/decoders represent the shallow-difference.
%。因为在训练中一次实现了多个域之间的转换，使得网络能更精准的表征不域与域间的深层共性，同时encoder 和decoder也能表示他们间分布的差异性.
Thanks to inner-similarity of face images which is captured by shared-latent space, it can be found that besides the change of hair color, our method maintains the original quality of human face and facial identity. More importantly, our results are obtained by training the network only once. While for three different kinds of hair, the UNIT and CycleGAN need to be trained three times.

%不难发现我们结果在头发颜色转换中取得了跟其他两个结果相比也有竞争性。经过仔细观察，可以发现除了头发颜色发生变化之外，我们的方法对人面部的造成的破环最少，尽可能的不让面部发生颜色或者形状的改变。效果上优于优于其他两个方法。更重要的是，我们是经过一次训练产生的domain 间的任意translation。对于3中不同的颜色来说，其他两个方法各需要3次end-to-end training。这进一步证明了我们方法的优越性。

\noindent \textbf{Painting style translation.} We further utilize the landscape photographs downloaded
from Flickr and WikiArt, which is also used in \cite{zhu2017unpaired}. The size of the dataset for each artist/style is 526, 1073, 400, and 563 for Cezanne, Monet, Van Gogh, and Ukiyo-e. In this experiment, we choose three of them, while Van Gogh is for incremental learning. Figure \ref{fig:painting} shows our results in comparison with the other methods. Compared with UNIT, our results are superior than it. Particularly, Our generated images are more clear and with higher contrast ratio than UNIT's, whilst comparable to CycleGAN's.
%从图像质量上看，我们生成的图像相比他更清晰，对比度更强。因为在训练中一次实现了多个域之间的转换，使得网络能更精准的表征不域与域间的深层共性，同时encoder和decoder也能表示他们间分布的差异性. 在这种情况下，我们这是我们比他们好的根源。

\begin{figure}
	\vspace{0pt}
	%\resizebox{0.48\textwidth}{!}{ %
	\begin{center}
		\setlength{\fboxrule}{0pt}
		\fbox{\includegraphics[width=0.47\textwidth]{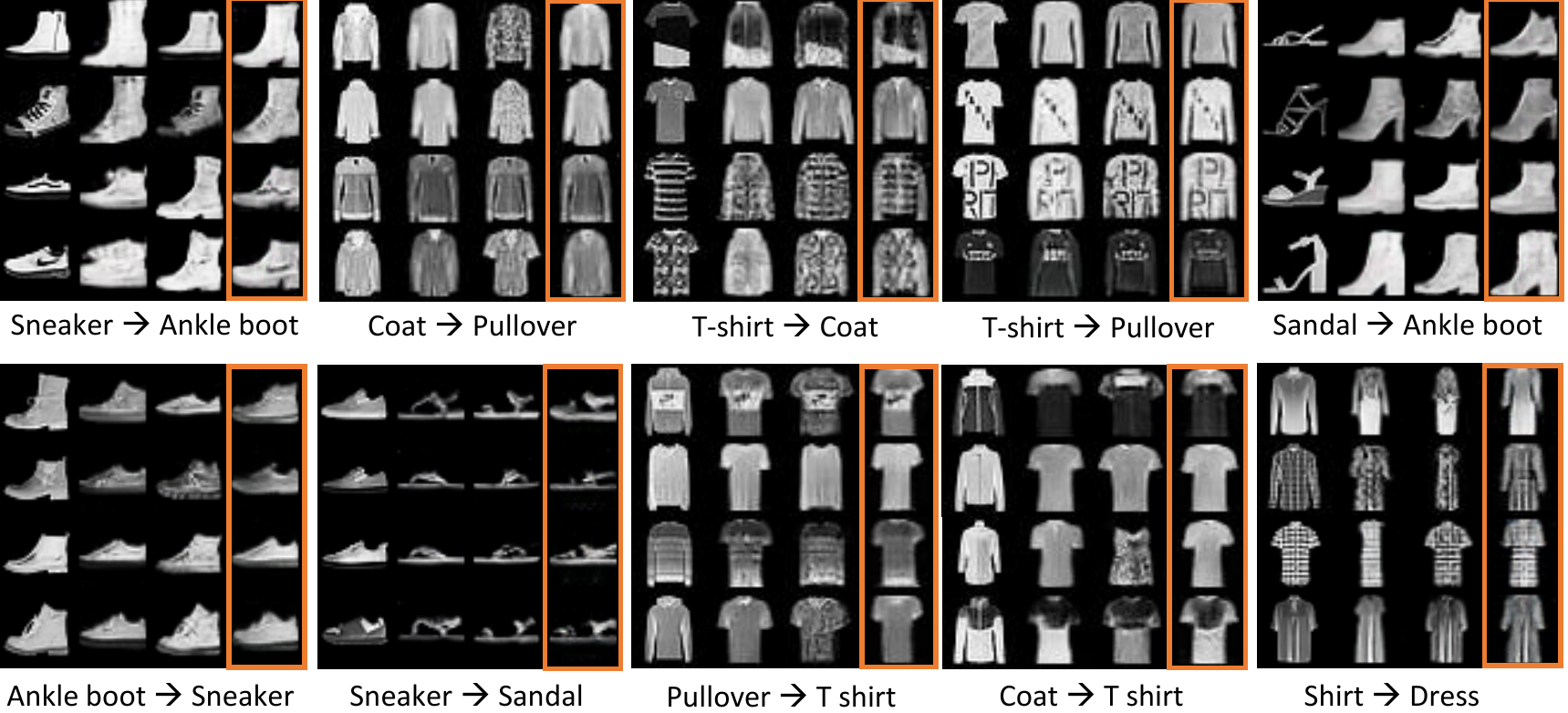}}
	\end{center}
	%}
	\vspace{-10pt}
	\caption{The translation between different categories of clothes. Here are ten groups of results, and each group has four columns. The first column is the input image. The second and third columns are generated by UNIT and CycleGAN. The generated images of our method are marked with \textcolor[rgb]{1, 0.3, 0.01} {orange} boxes. In the texture and details of the generated images, our method better preserves the texture information and is clearer the other two methods in details. In addition, the results we have obtained through end-to-end training only once.}
	\label{fig:fmnist}
	\vspace{-10pt}
\end{figure}

\noindent \textbf{Fashion-clothes translation.} We shows several example results achieved on a recently released dataset Fashion-MNIST \cite{xiao2017fashion}, which contains roughly 5 domains of clothes and 3 domains of shoes. Precisely, clothes are composed of T-Shirt, Pullover, Coat, Shirt and Dress, whilst shoes include Sandals, Sneaker and Ankle boots. We treat different categories of clothes as different domains.
Figure \ref{fig:fmnist} shows several results of translation between different clothes. In the texture and details of generated images, our method keeps the original texture of the clothes and have clearer clothing details. In general, our method obtains superior translation results to others.
\begin{figure*}
	\vspace{-25pt}
	\begin{center}
		\setlength{\fboxrule}{0pt}
		\fbox{\includegraphics[width=\textwidth]{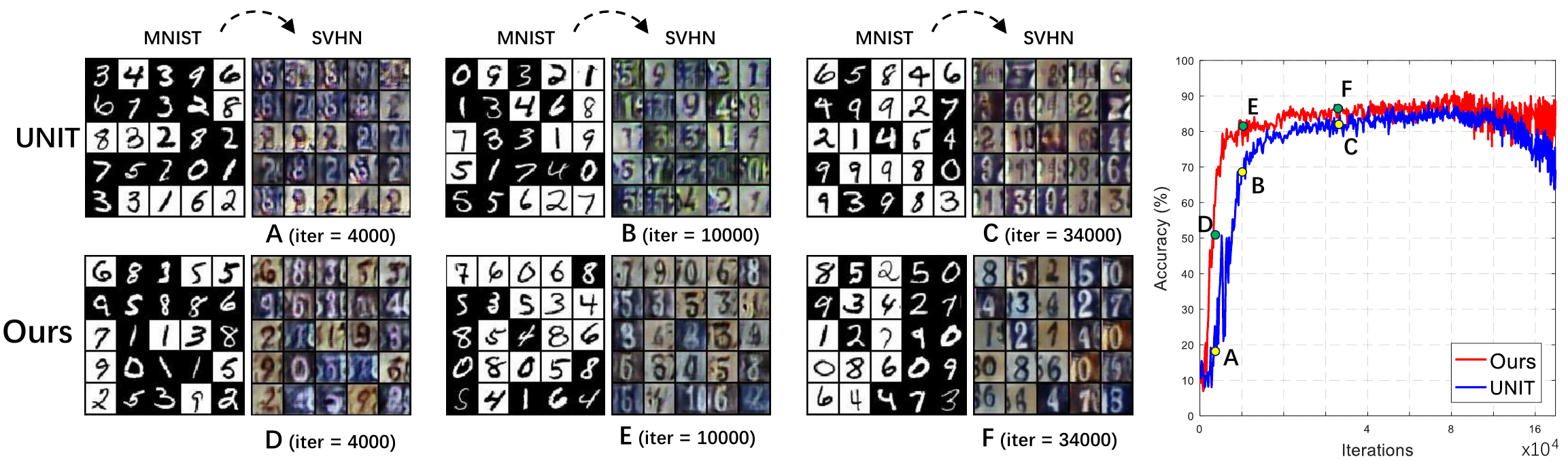}}
	\end{center}
	\vspace{-10pt}
	\caption{The visualization of digital image translation. The curve on the right represents the classification accuracy of MNIST (In the case of test, we classify the MNIST dataset using the features extracted by the discriminator in the SVHN dataset.), and the six images numbered from {\bf A} to {\bf F} are the visualization of different iterations.  Visibly, our results outperform in terms of image quality and details for describing digits. Due to the details for digits, our results obtain higher accuracy for classification by unsupervised learning without a hitch. Worth mentioning, our approach speed up the train process.}
	\label{fig:digit}
	\vspace{-13pt}
\end{figure*}

\vspace{5pt}
\noindent \textbf{Summary of Complexity Comparisons.} To demonstrate the advantages of complexity of our proposed framework, we compare the training time and model parameters with those baselines in Table \ref{tab:compare}. It can be clearly seen that our \emph{Domain-Bank} has less parameters and training time. This is because others are able to capture only one pair specific domains translation, which leads to an unavoidable burden under the situation where $n \times (n - 1)$ transformations are needed, given $n$ domains. However, in our framework, we merely require end-to-end training once where we can efficiently accomplish the translation between arbitrary two domains.
\begin{table}[h]
	\begin{center}
		\vspace{3pt}
		\resizebox{0.47\textwidth}{!}{ %
			%\begin{tabular}{c|c|c|c|c}
			%	\hline
			%	\small{Experiment} & \small{Type} & \small{UNIT \cite{liu2017unsupervised}} & \small{CycleGAN \cite{zhu2017unpaired}} & {\bf ours} \\
			%	\hline
			%	\hline
			%	\multirow{2}*{Face (3)} & Time & 1.86s$\times$3 & 3.98s$\times$3 & {\bf 3.13s}\\
			%	\cline{2-5}
			%	& Param & 18.02M$\times$3 & 22.76M$\times$3 & {\bf 25.58M} \\
			%	\hline
			%	\multirow{2}*{Clothes (5)} & Time & 1.79s$\times$10 & 3.98s$\times$10 & {\bf 5.95}\\
			%	\cline{2-5}
			%	& Param & 3.27M$\times$10  & 22.76M$\times$10 & {\bf 7.28M}\\
			%	\hline
			%	\multirow{2}*{Painting (4)} & Time & 1.79s$\times$6 & 3.98s$\times$6 & {\bf 3.32s}\\
			%	\cline{2-5}
			%	& Param & 3.27M$\times$6 & 22.76M$\times$6 & {\bf 5.94M}\\
			%	\hline
			%\end{tabular}
			\begin{tabular}{c|c|c|c|c}
				\hline
				\small{Experiment} & \small{Type} & \small{UNIT \cite{liu2017unsupervised}} & \small{CycleGAN \cite{zhu2017unpaired}} & {\bf ours} \\
				\hline
				\hline
				%\multirow{2}*{Face (3)} & Time & 5.58s & 11.94s %& {\bf 3.13s}\\
				%\cline{2-5}
				%& Param & 54.06M & 68.28M & {\bf 25.58M} \\
				%\hline
				%\multirow{2}*{Painting (4)} & Time & 10.74s & %23.88s & {\bf 3.32s}\\
				%\cline{2-5}
				%& Param & 19.62 & 136.56M & {\bf 5.94M}\\
				%\hline
				%\multirow{2}*{Clothes (5)} & Time & 17.92s & %39.85s & {\bf 5.95s}\\
				%\cline{2-5}
				%& Param & 32.75M & 227.65M & {\bf 7.28M}\\
				\multirow{2}*{Face (3)} & Time & 6 day & 13 day & {\bf 3 day}\\
				\cline{2-5}
				& Param & 54.06M & 68.28M & {\bf 25.58M} \\
				\hline
				\multirow{2}*{Painting (4)} & Time & 12 day & 27 day & {\bf 4 day}\\
				\cline{2-5}
				& Param & 19.62 & 136.56M & {\bf 5.94M}\\
				\hline
				\multirow{2}*{Clothes (5)} & Time & 21 day & 46 day & {\bf 7 day}\\
				\cline{2-5}
				& Param & 32.75M & 227.65M & {\bf 7.28M}\\
				\hline
			\end{tabular}
		}
	\end{center}
	\vspace{-7pt}
	\caption{The cost of parameters and times in the process of training. Obviously, advantages of our method in retrenching calculating space and time is revealed in this form.}
	\vspace{-15pt}
	\label{tab:compare}
\end{table}

\subsection{Quantitative Performance}
In order to better understand the performance gained by sharing more information through more than two domains, we adopt our framework to the domain adaptation task, which adapts a classifier trained using labeled samples in one domain (source domain) to classify samples in a new domain where labeled samples in the new domain (target domain) are unavailable during training. In our case, we append additional auxiliary domains by applying our framework with minimal efforts to check whether it can boost the system's performance.

More specifically, we utilize three datasets for digits: the Street View House Number (SVHN) dataset \cite{netzer2011reading}, the MNIST dataset \cite{lecun1998gradient} and USPS dataset \cite{friedman2001elements}, and perform multi-task learning where our framework is supposed to 1) translate images between any two of three domains and 2) classify the samples in the source domain using the features extracted by the discriminator in it. In the practice, we adopt a small network because the digit images have a small resolution.
\begin{table}[h]
	\centering
	\vspace{-7pt}
	\begin{tabular}{c|c|c|c}
		%\toprule
		\hline
		{Method} & \small{CoGAN \cite{liu2016coupled}}  & \small{UNIT \cite{liu2017unsupervised}} & {\bf ours}\\
		%\midrule
		\hline
		\hline
		\small{SVHN $\rightarrow$ MNIST} & \small{-} & \small{0.9053\%} & \small{{\bf 0.9146\%}} \\
		%\midrule
		\hline
		\small{MNIST $\rightarrow$ USPS} & \small{0.9565\%} & \small{0.9597\%} & \small{{\bf 0.9645\%}} \\
		%\midrule
		\hline
		\small{USPS $\rightarrow$ MNIST} & \small{0.9315\%} & \small{0.9358\%} & \small{{\bf 0.9412\%}} \\
		\hline
		%\bottomrule
	\end{tabular}
	\vspace{+3pt}
	\caption{Unsupervised domain adaption performance. The reported numbers are classification accuracies.}
	\vspace{-5pt}
	\label{tab:digit_translation}
\end{table}
In the experiment, we find that the cycle-consistency constraint is not necessary for this problem, and that is why we remove the cycle-consistency stream from the framework. In addition, we also tie the weights of the high-level layer of $D_a, a \in [1, n]$ in order to adapt a classifier trained in the source domain to the target domain.

As a result, Figure \ref{fig:digit} shows the visualization of digit and Table \ref{tab:digit_translation} reports the achieved performance with comparison to the competing algorithms. We achieve better performance for SVHN $\rightarrow$ MNIST task than the UNIT approach, which is the state-of-the-art right now. We also obtain the superior results than UNIT on MNIST $\rightarrow$ USPS and USPS $\rightarrow$ MNIST tasks.
\vspace{-5pt}
\section{Capabilities of Our Framework}
\label{sec_capability}
\subsection{Incremental Training}
Retraining a new model is inconvenient and risks being not able to recover the performance of previous trained model, when a new style needs to be added. Our framework proposed in this paper has same capability described in \cite{chen2017stylebank} supporting incremental training.

\begin{figure}[h]
	\vspace{-10pt}
	\begin{center}
		\setlength{\fboxrule}{0pt}
		\fbox{\includegraphics[width=0.47\textwidth]{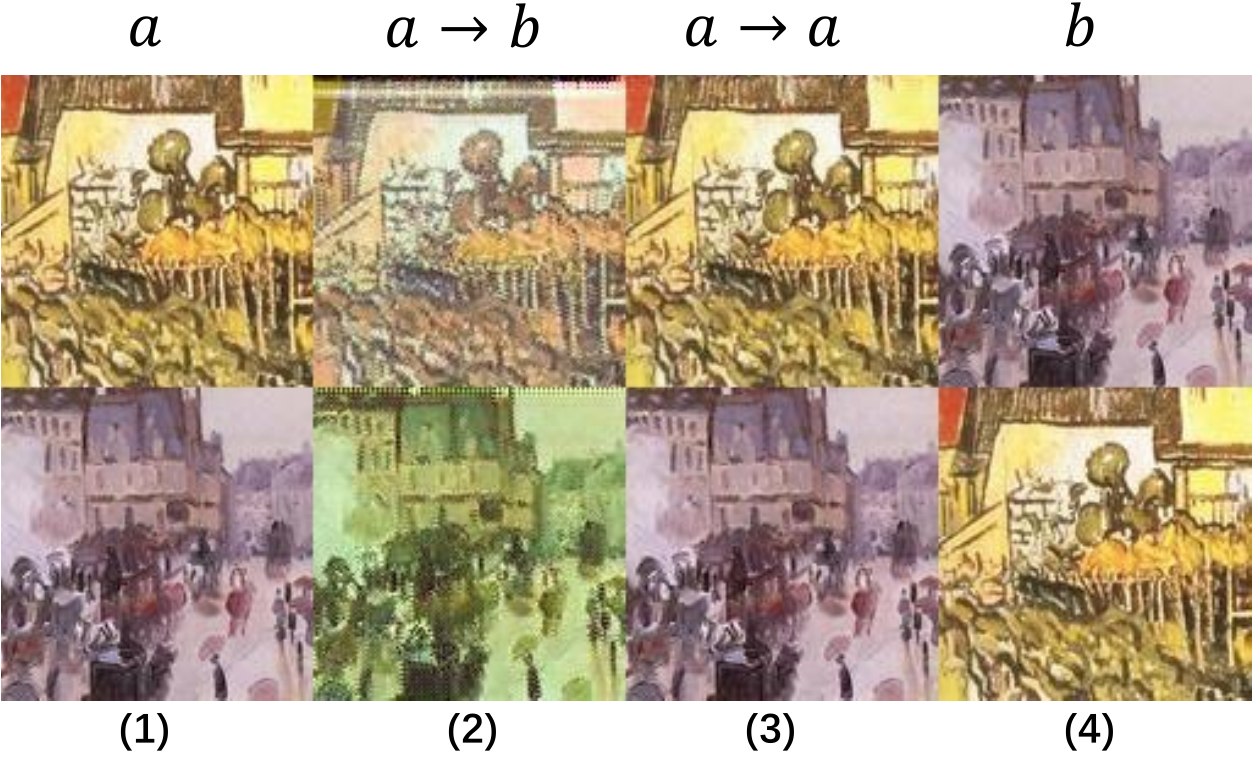}}
	\end{center}
	\vspace{-10pt}
	\caption{Results of translation between incrementally trained Painting style of Van Gogh  and other well-trained domain are shown in this figure. The images on far left column are inputs of each row, and the second column the translated images. The third column are reconstructed images for the corresponding input images, whilst the last column are images from target domain.}
	\vspace{-7pt}
	\label{fig:incremental_learning}
\end{figure}

\begin{figure}[h]
	\begin{center}
		\setlength{\fboxrule}{0pt}
		\fbox{\includegraphics[width=0.47\textwidth]{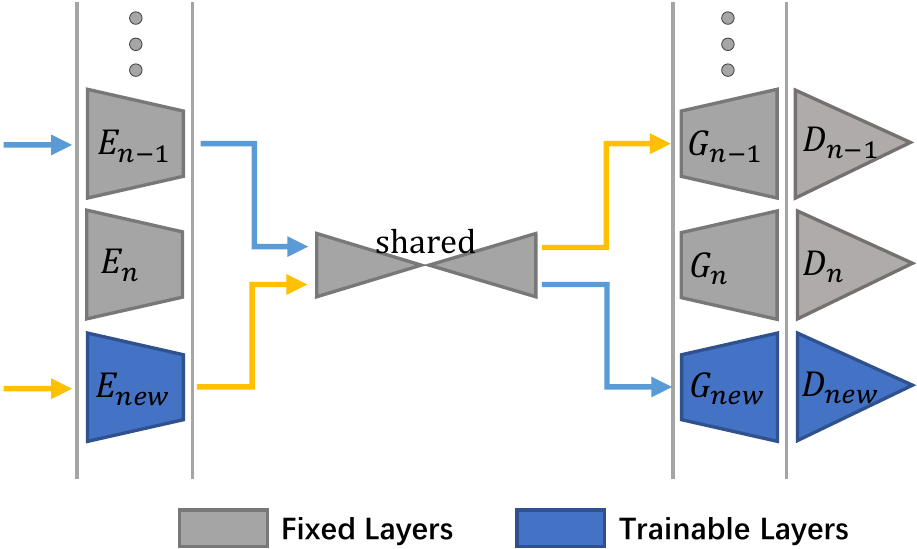}}
	\end{center}
	\vspace{-10pt}
	\caption{The blue parts are encoder/decoder and discriminator front layers of incremental domain, which are trainable in incremental training process, whereas the gray parts are for other domains well trained which are fixed.
	}
	\vspace{-10pt}
	\label{fig:incremental_learning_network}
\end{figure}

When we need add a new style, only the images sampled from the specific incremental domain participate in the training process. The incremental domain's encoder/decoder and discriminator layers are not fixed shown in Figure \ref{fig:incremental_learning_network}. Considering the incremental domain $X_c$ and an existing domain $X_a$, a's encoder/decoder will be fixed and only samples
$\{x_i^c\}_{i=1}^{N_c}$ where $\{x_i^c \in X_c\}$ assist incremental training. The loss function for incremental training is defined as follow ($n$ is the number of domain):
\vspace{-3pt}
\begin{eqnarray}
\mathcal{L}(E, G, D)=\sum_{j}^{n} \{{\mathcal{L}}_{VAE_c}(E_c, G_c) \nonumber \\
+ \mathcal{L}_{GAN_{cj}}(E_c, G_j, D_j) \nonumber \\
+ \mathcal{L}_{cyc_{cj}}(E_c, G_j, E_j, G_c)\}
\end{eqnarray}

Furthermore, we show few samples in Figure \ref{fig:incremental_learning} to demonstrate the efficiency.

\begin{figure}[h]
	\begin{center}
		\setlength{\fboxrule}{0pt}
		\fbox{\includegraphics[width=0.47\textwidth]{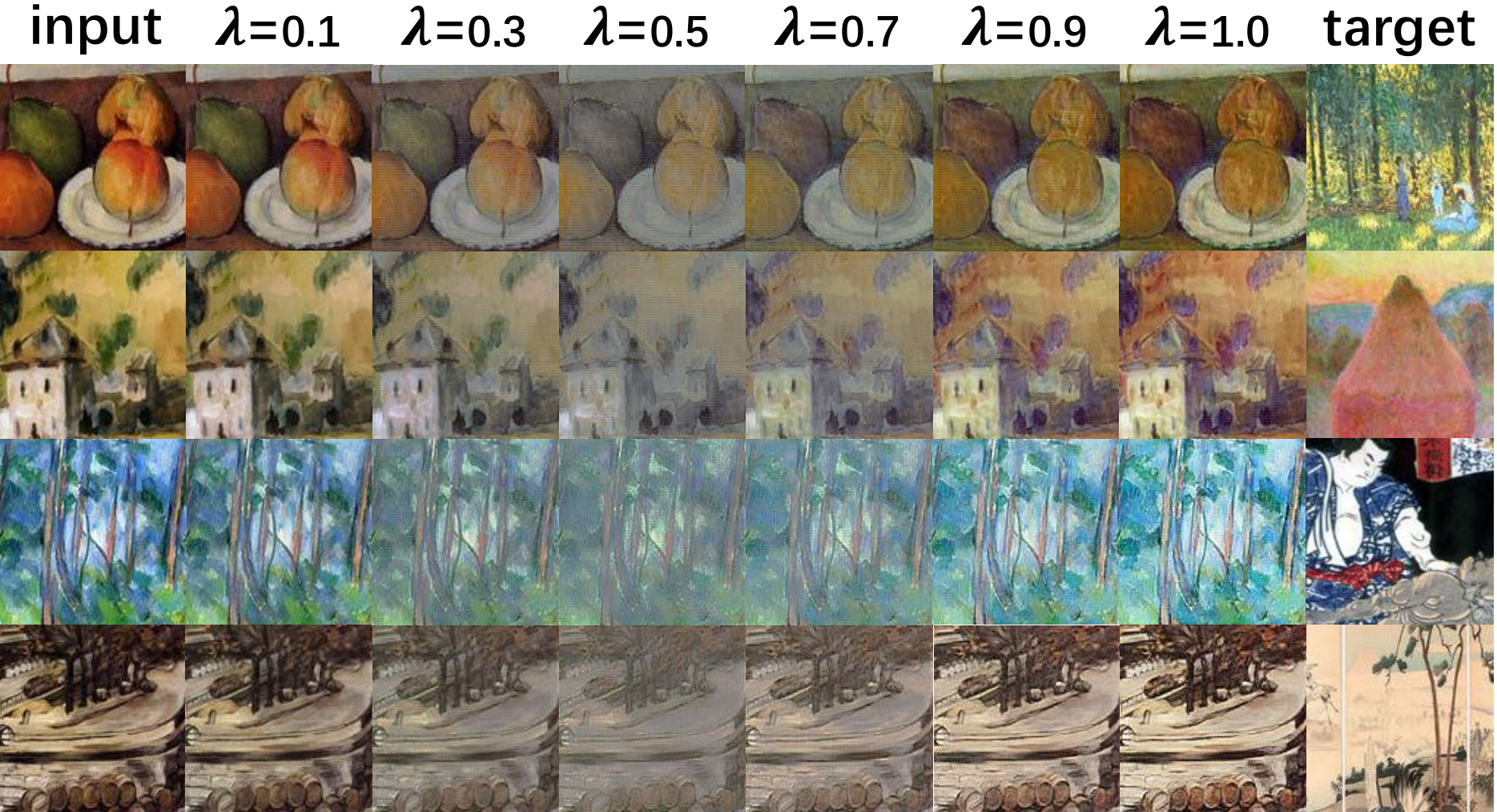}}
	\end{center}
	\vspace{-10pt}
	\caption{Results of fusion of two styles with variant ratio $\lambda$ are provided in this figure and on the far left column are input images, while the far right are target domain images. Every row shows progressive translation between two styles.}
	\vspace{-10pt}
	\label{fig:fusion}
\end{figure}

\subsection{Domain Fusion}
In this section, we demonstrate an experiment for style fusion: linear fusion of two different styles.
\noindent \textbf{Style Linear Fusion.} Translations between different styles are encoded into different pairs of $\{E_{source}, G_{target}\}$, especially, $E_{source}$ for the input port for source domain and $G_{target}$ for the output port for target domain. We linearly combine $G_{target}$ for different target domains in Domain-Bank layers and the fused $G^*$ can be a new output port:
\vspace{-5pt}
$$G^* = \lambda*G_1 + (1-\lambda)*G_2~~~~~~~~~~\lambda\in(0,1)$$
Figure \ref{fig:fusion} shows progressive results of two styles with variant ratio $\lambda$.

\section{Conclusion}
\label{sec_conclusion}
In this paper, we have proposed a novel multi-domain image translation framework, namely \emph{Domain-Bank}. We show it learnt to translate from multiple domains to multiple domains in one training process. Particularly, our \emph{Domain-Bank} explicitly reduces the training time from $O(n^2)$ to $O(n)$, given $n$ domains. The universal shared auto-encoder subnetwork  leads to a better generation which is confirmed in both quantitative and qualitative experimental results. More notably, our framework has less parameters and training time when comparing to others. In addition, we also provide more functional augmentations like domain linear combination and incremental learning.

{\small
\bibliographystyle{ieee}
\bibliography{egbib}
}

\end{document}